\documentclass{article}

\PassOptionsToPackage{numbers, compress}{natbib}

\usepackage[final]{nips_2018}
\usepackage{graphicx,wrapfig,lipsum}
\usepackage{multicol}
\usepackage[export]{adjustbox}

\usepackage[utf8]{inputenc} 
\usepackage[T1]{fontenc}    
\usepackage{hyperref}       
\usepackage{url}            
\usepackage{booktabs}       
\usepackage{amsfonts}       
\usepackage{nicefrac}       
\usepackage{microtype}      

\newcommand{\ggold}{\mathcal{G}_{\textrm{Gold}}}

\newcommand{\gadd}{\mathcal{G}_{\textrm{Add}}}
\newcommand{\gnoise}{\mathcal{G}_{\textrm{Noise}}}

\newcommand{\beginsupplement}{%
        \setcounter{table}{0}
        \renewcommand{\thetable}{S\arabic{table}}%
        \setcounter{figure}{0}
        \renewcommand{\thefigure}{S\arabic{figure}}%
        \setcounter{section}{0}
        \renewcommand{\thesection}{Supplementary Material}
        \setcounter{subsection}{0}
        \renewcommand{\thesubsection}{S\arabic{subsection}}%
        \setcounter{equation}{0}
     }

\usepackage{amsmath}
\usepackage{graphicx}
\usepackage{caption}
\usepackage{threeparttable}
\usepackage{ragged2e}
\usepackage{subcaption}
\usepackage{mathtools}
\usepackage{booktabs}
\usepackage{todonotes}
\usepackage{subfiles}
\usepackage{titlesec}
\usepackage[export]{adjustbox}

\title{Interpretable Graph Convolutional Neural Networks for Inference on Noisy Knowledge Graphs}

\author{
  Daniel Neil \quad Joss Briody \quad Alix Lacoste \quad Aaron Sim \quad Paidi Creed \quad Amir Saffari\\
  BenevolentAI \\
  Brooklyn, NY and London, UK\\
  \texttt{\{daniel.neil,joss.briody,alix.lacoste},\\
  \texttt{aaron.sim,paidi.creed,amir.saffari\}@benevolent.ai}
}

\begin{document}

\maketitle

\begin{abstract}
In this work, we provide a new formulation for Graph Convolutional Neural Networks (GCNNs) for link prediction on graph data that addresses common challenges for biomedical knowledge graphs (KGs).  We introduce a regularized attention mechanism to GCNNs that not only improves performance on clean datasets, but also favorably accommodates noise in KGs, a pervasive issue in real-world applications. Further, we explore new visualization methods for interpretable modelling and to illustrate how the learned representation can be exploited to automate dataset denoising. The results are demonstrated on a synthetic dataset, the common benchmark dataset FB15k-237, and a large biomedical knowledge graph derived from a combination of noisy and clean data sources. Using these improvements, we visualize a learned model's representation of the disease cystic fibrosis and demonstrate how to interrogate a neural network to show the potential of PPARG as a candidate therapeutic target for rheumatoid arthritis.
\end{abstract}

\section{Introduction and Motivation}
In biomedicine, knowledge graphs are critical in understanding complex diseases and advancing drug discovery. The problem of identifying novel therapeutic targets for a given disease for example can be formulated as a link prediction problem, a major area of study in statistical relational learning \cite{getoor2007introduction}. Various tensor factorization methods have found success
\cite{nickel2011three,bordes2013translating,yang2014embedding,trouillon2016complex}, as have convolutional neural networks (CNNs) \cite{toutanova2015representing,dettmers2017convolutional,nguyen2017novel} and path-based methods \cite{xiong2017deeppath,das2017go}.
Recently, graph convolutional neural networks (GCNNs) have been applied to this problem, outperforming a number of standard models \cite{schlichtkrull2017modeling}.

Real-world knowledge graphs tend to contain relationships from multiple sources of varying quality. For example, drug-target associations extracted from unstructured text  are less reliable than manually curated ones. In this work, we make two important contributions. First, we demonstrate that introducing a learnable link weight outperforms existing tensor factorization and GCNN models in the presence of noise. This new model assigns low weights to unreliable edges, which can be viewed as learning an edge filter to remove unreliable or uninformative edges from the knowledge graph. Second, we demonstrate that this model is more interpretable because it allows measuring the impact of a particular edge on a prediction by adjusting the link weight or removing the edge entirely. Moreover, when our knowledge graph is constructed by combining information from a number of diverse sources of variable reliability, the learned link weights can be used to assess the quality or relevance of different data sources. These benefits are illustrated with applications in drug-target discovery, where the added value of interpretability is particularly great.

\section{Model Formulation}
Let $\mathcal{E}$ denote the set of all entities and $\mathcal{R}$ the set of all relation types in a KG, represented by a directed multigraph $\mathcal{G}$. An element of the KG can be represented by the triple $(e_s,r,e_o) \in \mathcal{E}\times\mathcal{R}\times\mathcal{E}$. Knowledge graph embedding models learn vector representations of entities, $\textbf{e}_i\in\mathbb{R}^{d_e}$, as well as relations $\textbf{r}\in\mathbb{R}^{d_r}$, and a mapping (decoder) $f: \mathbb{R}^{d_e}\times\mathbb{R}^{d_r}\times\mathbb{R}^{d_e} \rightarrow [0,1]$ assigning a probability of existence to each triple.
To learn the entity embeddings, we use a GCNN model. Our implementation re-writes the GCNN introduced in \cite{kipf2016semi,zitnik2018modeling} as a series of matrix and element-wise multiplications:
\begin{align} \label{eq:finalgcnn}
H^{(l+1)} = \sigma \bigg(B^{(l)} + \sum_{r \in \mathcal{R}} (C_r \odot A_r) (H^{(l)} W_{r}^{(l)}) \bigg)
\end{align}
where $\sigma (\cdot)$ is an element-wise nonlinear function (e.g. a ReLU \cite{nair2010rectified}),
$H^{(0)} := I_N$ is the N-dimensional identity matrix, $B^{(l)} \in \mathbb{R}^{N \times k^{(l)}}$ is a $k^{(l)}$ dimensional bias for each node and $C_r \in \mathbb{R}^{N \times N}$ is a fixed scaling factor between two connected nodes. The adjacency matrix  is $A_r \in \{0, 1\}^{N \times N}$, the hidden representation $H^{(l)} \in \mathbb{R}^{N \times k^{(l)}}$, and the weight matrix is denoted, $W_r^{(l)} \in \mathbb{R}^{k^{(l+1)} \times k^{(l)}} \forall r \in \mathcal{R}$. The final layer $H^{(L)}$ contains as its rows the embedding vector $\textbf{e}_i$ for each entity, i.e $d_e=k^{(l)}$. Relation vectors are learned by look-up in a simple embedding matrix $R\in\mathbb{R}^{|\mathcal{R}|}\times\mathbb{R}^{d_r}$. To calculate probabilities, we can use any choice of the decoder $f(\cdot)$ \cite{yang2014embedding,trouillon2016complex,nickel2011three}. This work focuses on both the Complex decoder (\cite{trouillon2016complex}) as well as the DistMult model:
\begin{align} \label{dist}
f(e_s, R_r, e_o) = e_s^T R_r e_o
\end{align}


We propose an attention model in which each link has an independent, learnable weight designed to approximate the \textit{usefulness} of that link. Following the success of \cite{velivckovic2017graph}, we constrain our attention weights to have a fixed total ``budget'':
\begin{align}
    C_{r,i,j} = \frac{1}{\sum_{r' \in \mathcal{R}} \sum_{j' \in \mathcal{N}_i^r} |\hat{C}_{r', i, j'}|} |\hat{C}_{r,i,j}|
\end{align}
where $\hat{C}_{r,i,j}$ is initialized to 1, such that $C_{r,i,j} = 1 / |\mathcal{N}_i|$ for all $j$ at the start of training. This encourages the model to select only useful links that maximally aid in predicting true facts.




\section{Experiments}

 All hyper-parameters were optimized using the mean reciprocal rank (MRR) performance on the validation set. This work employs a single GCNN layer, with a diagonalized weight matrix $W_r$ and no non-linearity applied. We experimented with more layers and non-linear transforms but this did not improve performance. To train the model, we minimize the cross-entropy loss on the link class $\in \{0, 1\}$, and perform negative sampling following \cite{nickel2016review,trouillon2016complex,bordes2013translating}.
 We performed grid search on the number of negatives sampled for each positive, $n\in \{1,10,20,50\}$, as well as the embedding dimension $d\in\{50, 100, 200, 300\}$.
 Our reported results all use $n=10$ and $d=300$. For best performance, dropout with probability 0.5 was used on both the embeddings and on the links themselves. All embeddings were initialized with an $L_2$-norm of 1, so that all entities initially contribute equally in magnitude to the embedded representation at the start of training.

\subsection{Performance on FB15k-237, with Attention}
 Having established the formulation of a GCNN with scalar attention, we now examine the relationship between data volume and noise. Four different conditions were trained, with results presented in Table \ref{tab:bench-perf}. Our setup uses a dataset of a given size (``50\%''), adding in noisy edges (``Noised'') in equal volume to the remaining 50\%, and skipping training on these possibly noisy edges (``Skip'') while the edges remain in $A_{r,i,j}$. We see that GCNNs with attention consistently outperform those without attention, with MRR of $0.283 \pm 0.006$ and $0.272 \pm 0.001$, and hits@10 of $0.482 \pm 0.009$ and $0.475 \pm 0.004$ respectively, with the ``Noised'' and ``Skip'' condition of particular relevance to standard biomedical knowledge graphs.

 \begin{table*}[t!]
  \centering
  \begin{tabular}{l llll llll}
    \toprule
     & \multicolumn{4}{c}{Hits@10} & \multicolumn{4}{c}{MRR}\\
    \cmidrule(r){2-5} \cmidrule(r){6-9}
    Algorithm         & 100\% &  50\%& Skip & Noised  & 100\% & 50\% & Skip & Noised \\
    \midrule
    DistMult          & 43.2 & 20.2 &  N/A  & 20.6 & 23.9 & 8.69 & N/A & 8.93 \\
    ComplEx           & 44.1 & 24.1 &  N/A  & 24.3 & 25.9 & 10.9 & N/A  & 11.0 \\
    GCNN & 47.5 & 33.2 & 25.8 & 21.4 & 27.2 & 16.8 & 13.3 & 11.1 \\
    GCNN w/att & \textbf{48.2} & \textbf{34.7} & \textbf{34.0} & \textbf{35.6} & \textbf{28.3} & \textbf{18.5} & \textbf{18.8} & \textbf{19.1}\\
     R-GCN+ (\cite{schlichtkrull2017modeling}) & 41.7 & \hspace{2mm} - & \hspace{2mm} - & \hspace{2mm} - & 24.9 & \hspace{2mm} - & \hspace{2mm} - & \hspace{2mm} -\\
    \bottomrule
  \end{tabular}
  \caption{Performance on the FB15k-237 Dataset. Our results compare favorably with those reported in previous GCNN studies.}
\label{tab:bench-perf}
\end{table*}
\begin{figure}[h]
    \begin{minipage}{.47\textwidth}
     To further explore the sensitivity of the GCNN model to noise, we replicate the experimental setup of \cite{pujara2017sparsity-and-noise} on the benchmark standard FB15k-237 dataset \citep{toutanova2015observed,bordes2013translating}. The results presented in Fig.\ref{fig:prob1_6_1}, right, reinforce previous findings that our proposed attention mechanism makes GCNNs more robust to noise. With an entirely clean knowledge graph, the addition of attention yields a 7.5\% improvement in performance. When approximately 20-30\% of the input graph is noise, the difference is $\approx$ 25-33\%.
    \end{minipage}%
    \hspace{5mm}
    \begin{minipage}{0.47\textwidth}
        \centering
        \includegraphics[width=0.75\linewidth, ]{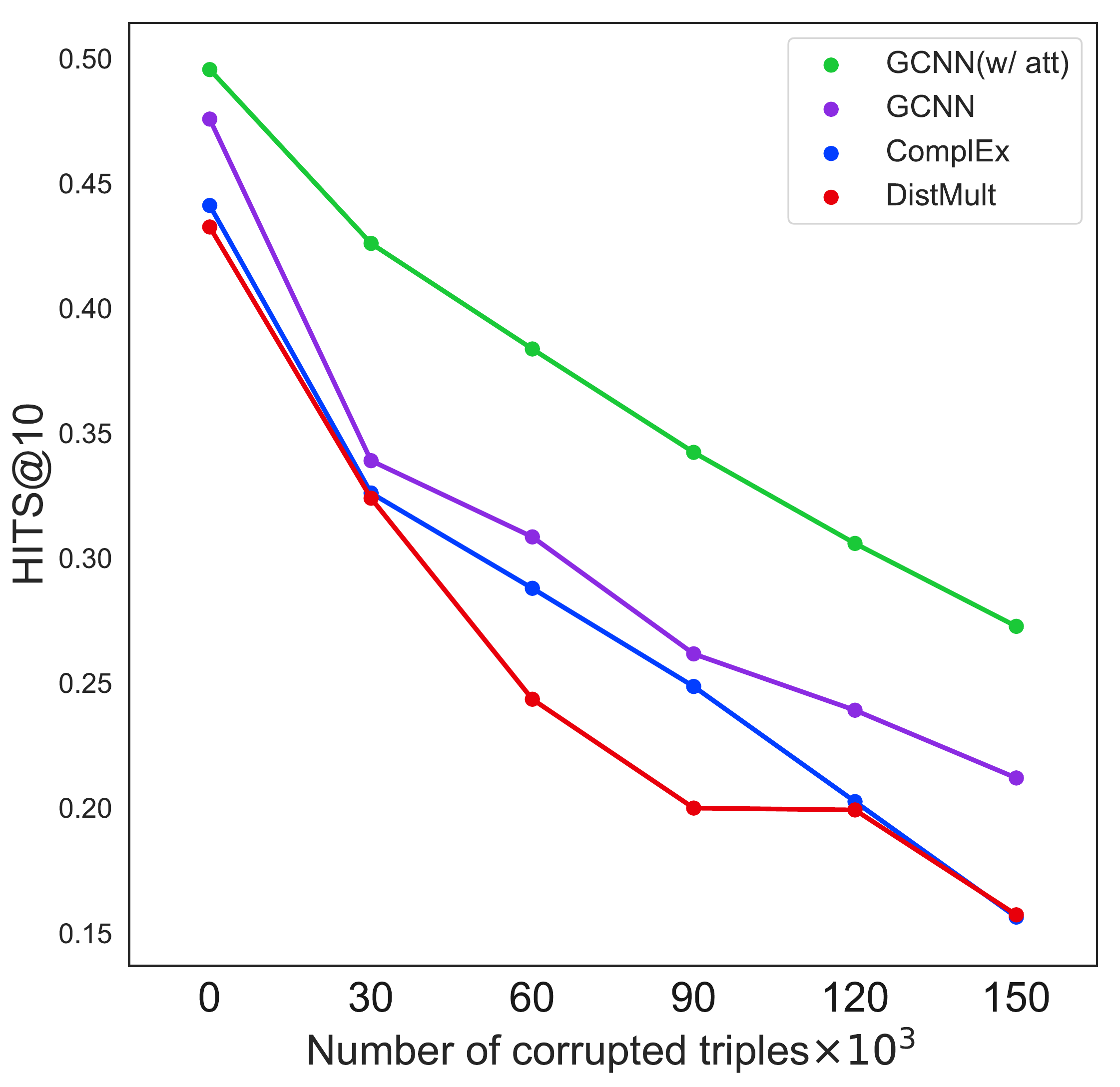}
        \caption{Test set HITS@10 for corrupted triples in FB15k-237.}
        \label{fig:prob1_6_1}
    \end{minipage}
\end{figure}



\subsection{Correctness and Interpretation in a Biological Knowledge Graph}

\begin{figure}[t!]
    \small
    \centering
    \begin{subfigure}[t]{0.27\textwidth}
        \includegraphics[width=\textwidth,]{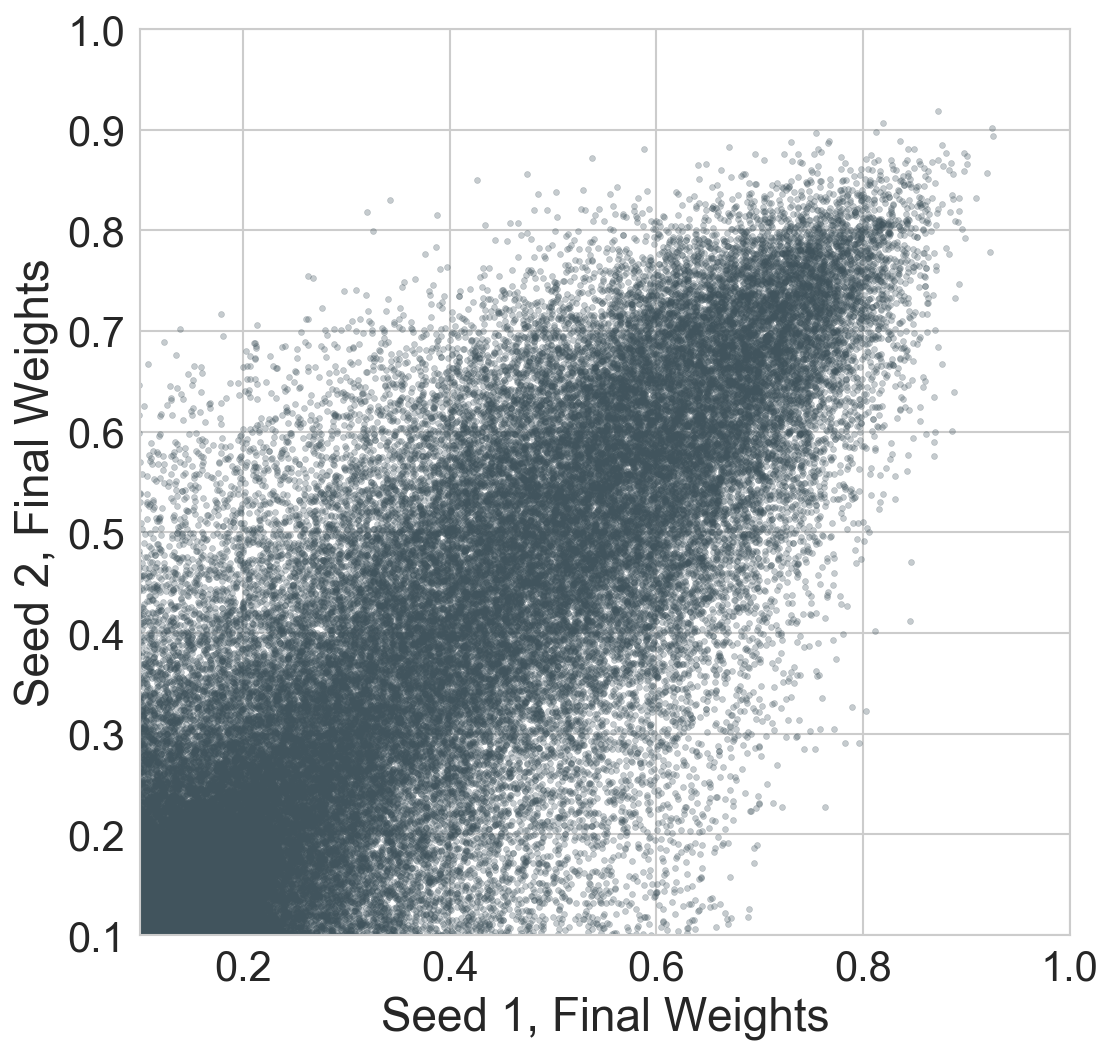}
        \label{fig:selfsim}
    \end{subfigure}
    ~
    \hspace{3mm}
    \begin{subfigure}[t]{0.25\textwidth}
        \includegraphics[width=\textwidth,]{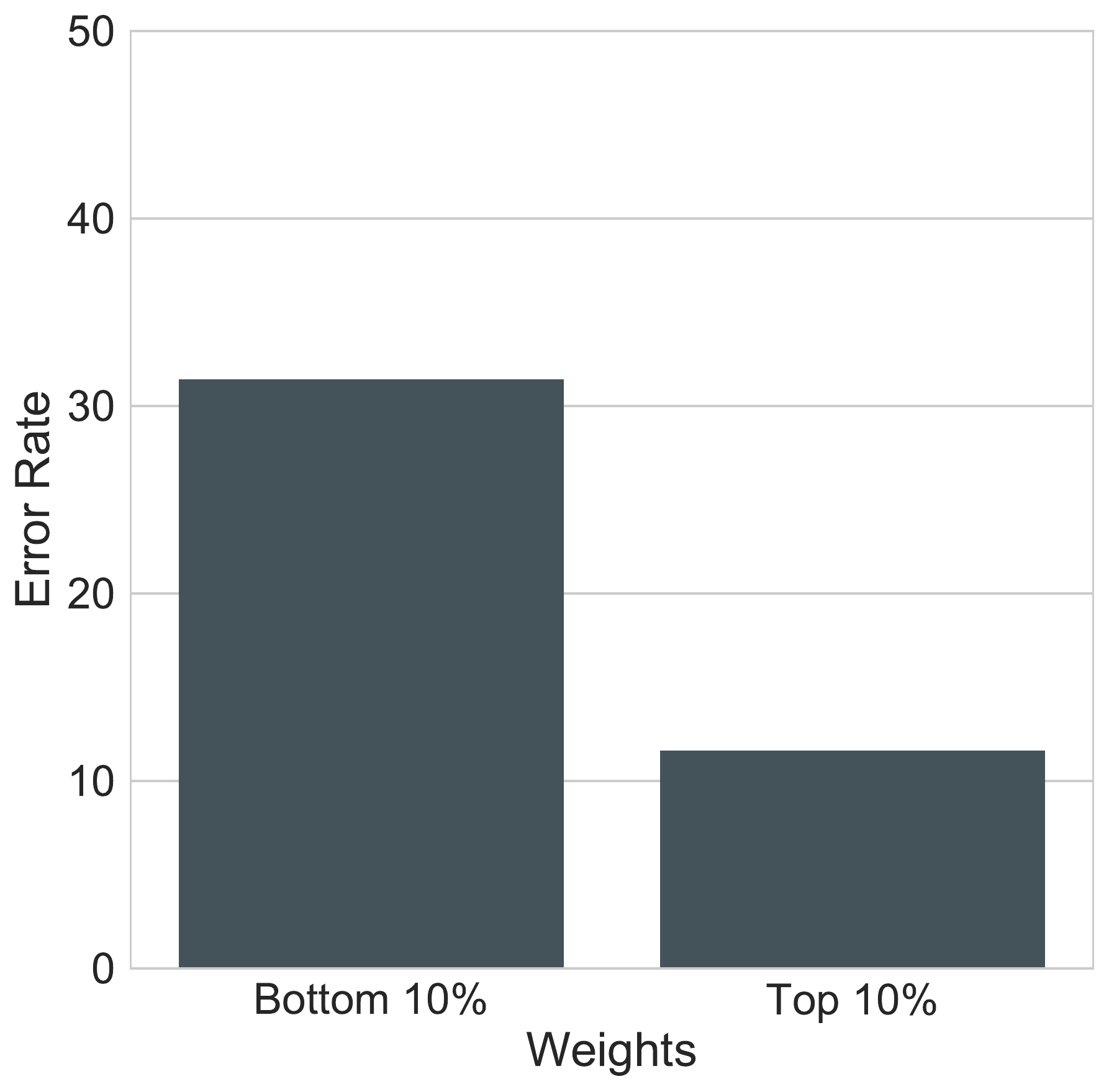}
        \label{fig:grounderrors}
    \end{subfigure}
    ~
    \hspace{3mm}
    \begin{subfigure}[t]{0.27\textwidth}
        \includegraphics[width=\textwidth,]{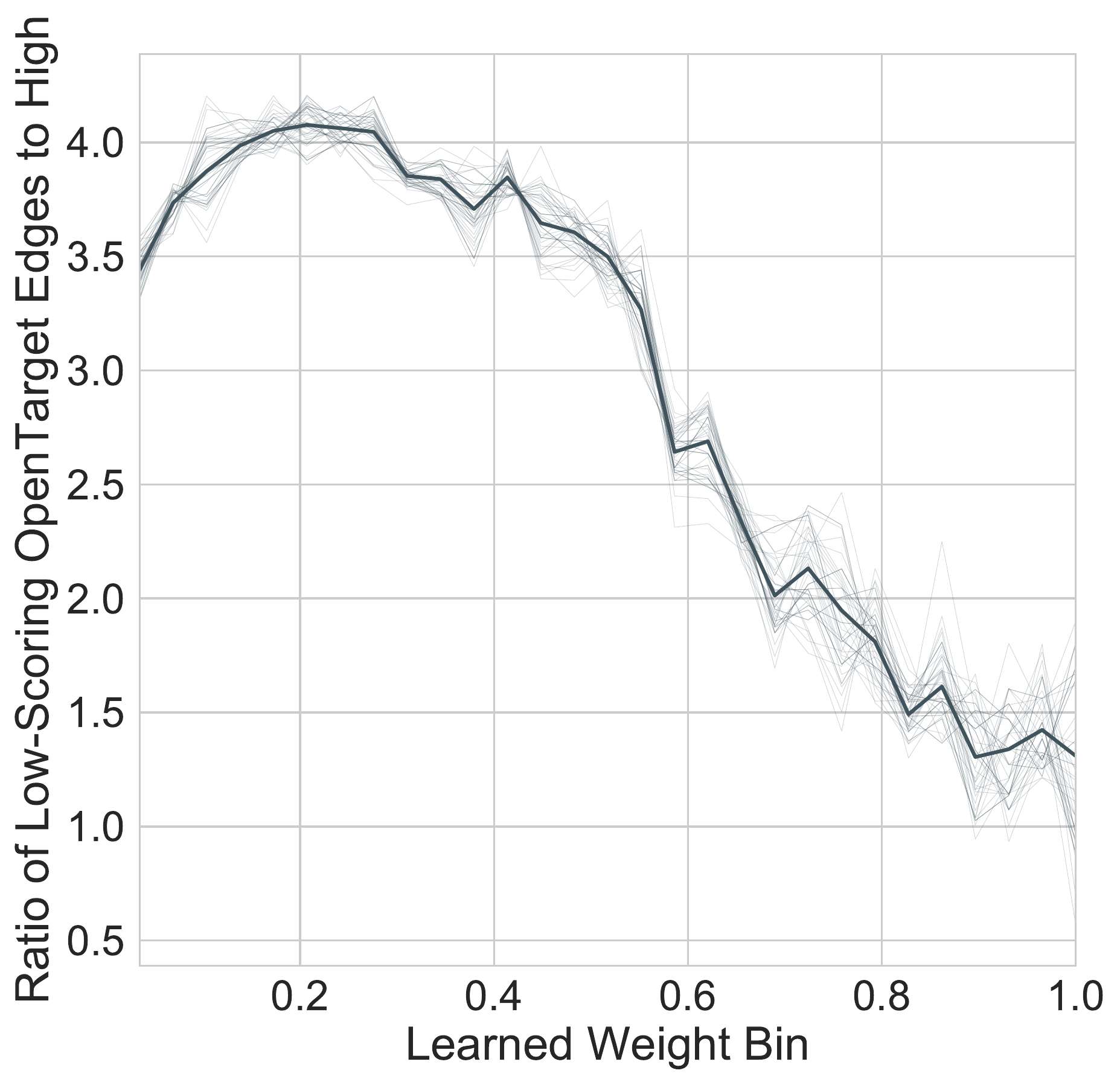}
        \label{fig:kgmeaning}
    \end{subfigure}
    ~
    \caption{
    \textbf{Left}: Self-similarity of weight edges from two random initializations on a drug, disease, and gene dataset.  Despite different initial conditions, most attention weights end training at similar magnitudes and thus lie along the diagonal.
    \textbf{Centre}: Rate of errors in grounding or relation extraction, when examining the top 10\% and bottom 10\% of weights.
    \textbf{Right}: Ratio of low-scoring Open Targets \citep{opentargets} edges to high; 5 runs with bootstrap sampling lines to show mean and variance.
    }
    \label{fig:correctness-all}
\end{figure}

As an example real-world application, we examine a subset of a proprietary knowledge graph with 708k edges, compiled from unstructured data sources including NCBI PubMed full text articles and a variety of structured sources including CTD \cite{Davis:2017fx}, KEGG \cite{Kanehisa2017}, OMIM \cite{OMIM}, BioGRID \cite{BioGRID}, Omnipath \cite{Turei:ki}, and ChEMBL \cite{Bento2014}.   Fig.~\ref{fig:correctness-all}, left, plots the weights of relations extracted from unstructured text between genes and diseases. It demonstrates that the weights have an inherent consistency despite uncorrelated random initial conditions (Pearson's $r=0.9$).  Further, this value can be examined as a measure of edge correctness. First, blinded manual evaluation of edges shows that low-weighted edges are three times more likely to be erroneous than high-weighted ones (Fig.~\ref{fig:correctness-all}centre). Second, edge weights are compared with their corresponding confidence scores in the Open Targets platform \cite{opentargets}, if present, and, remarkably, GCNN weights are predictive of this score.
As Fig.~\ref{fig:correctness-all}, right shows, a low-weighted edge, with score below 0.1, is 4 times more likely to be a low-scoring Open Targets edge than a high-weighted one with score above 0.9 ($p=6 \times 10^{-28}$, two-sided KS test).  This suggest that edge weight is indeed indicative of trustworthiness.

\begin{figure}[b!]
    \centering
    \begin{subfigure}[t]{0.27\textwidth}
        \includegraphics[width=\textwidth,]{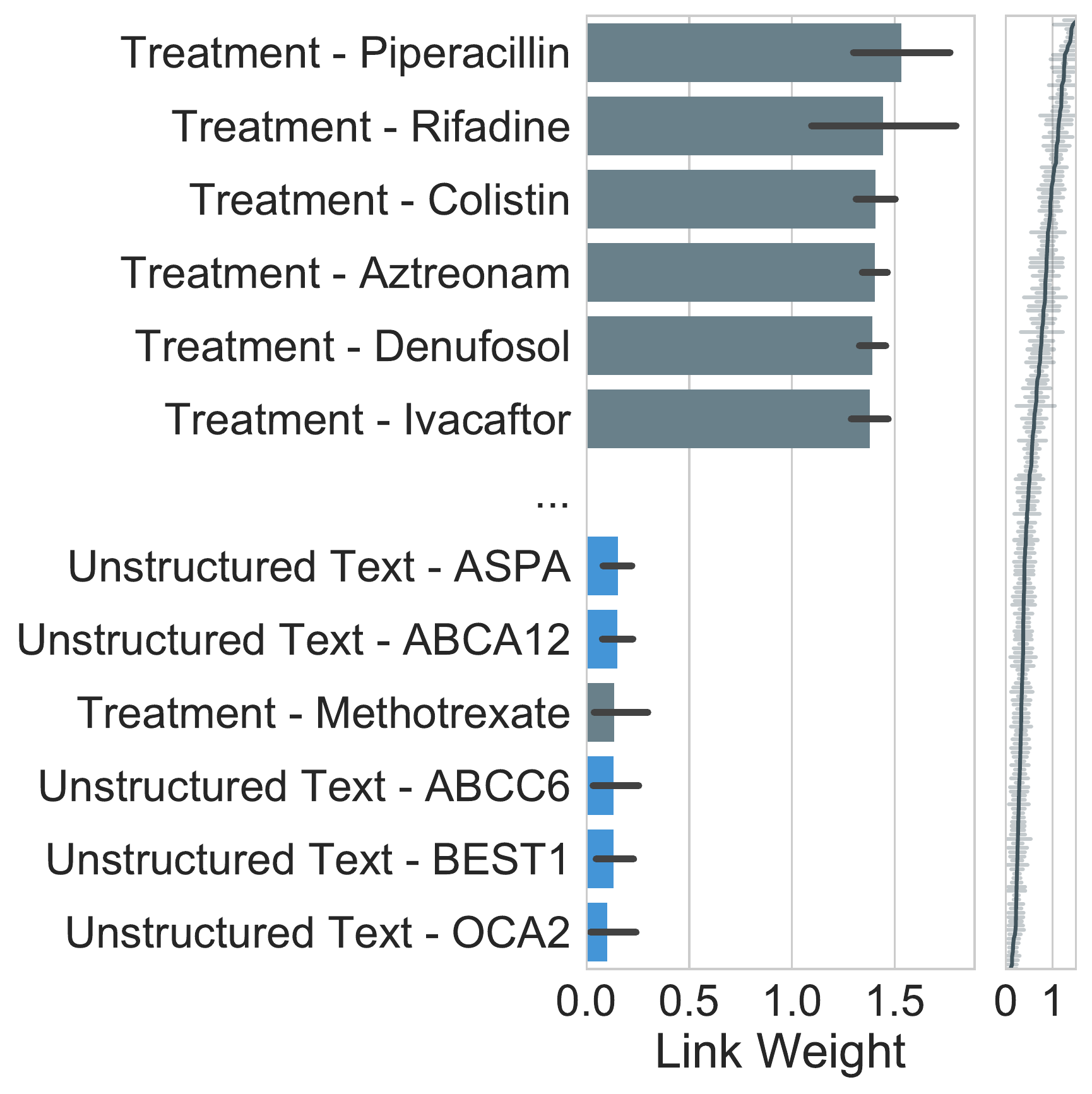}
        \label{fig:entityviz}
    \end{subfigure}
    ~
    \hspace{3mm}
    \begin{subfigure}[t]{0.25\textwidth}
        \includegraphics[width=\textwidth,]{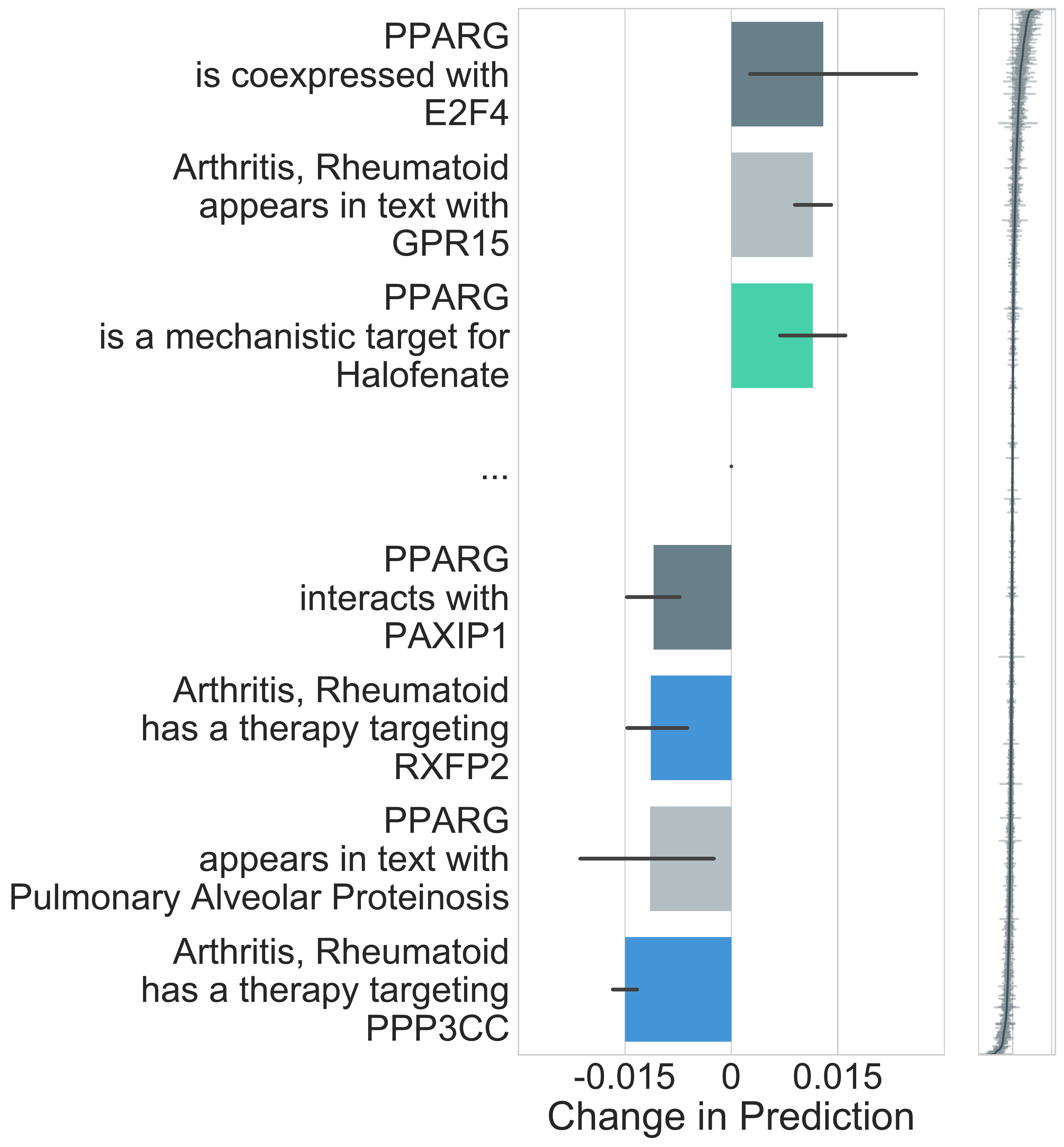}
        \label{fig:addremove}
    \end{subfigure}
    ~
     \hspace{4mm}
    \begin{subfigure}[t]{0.27\textwidth}
        \includegraphics[width=\textwidth,]{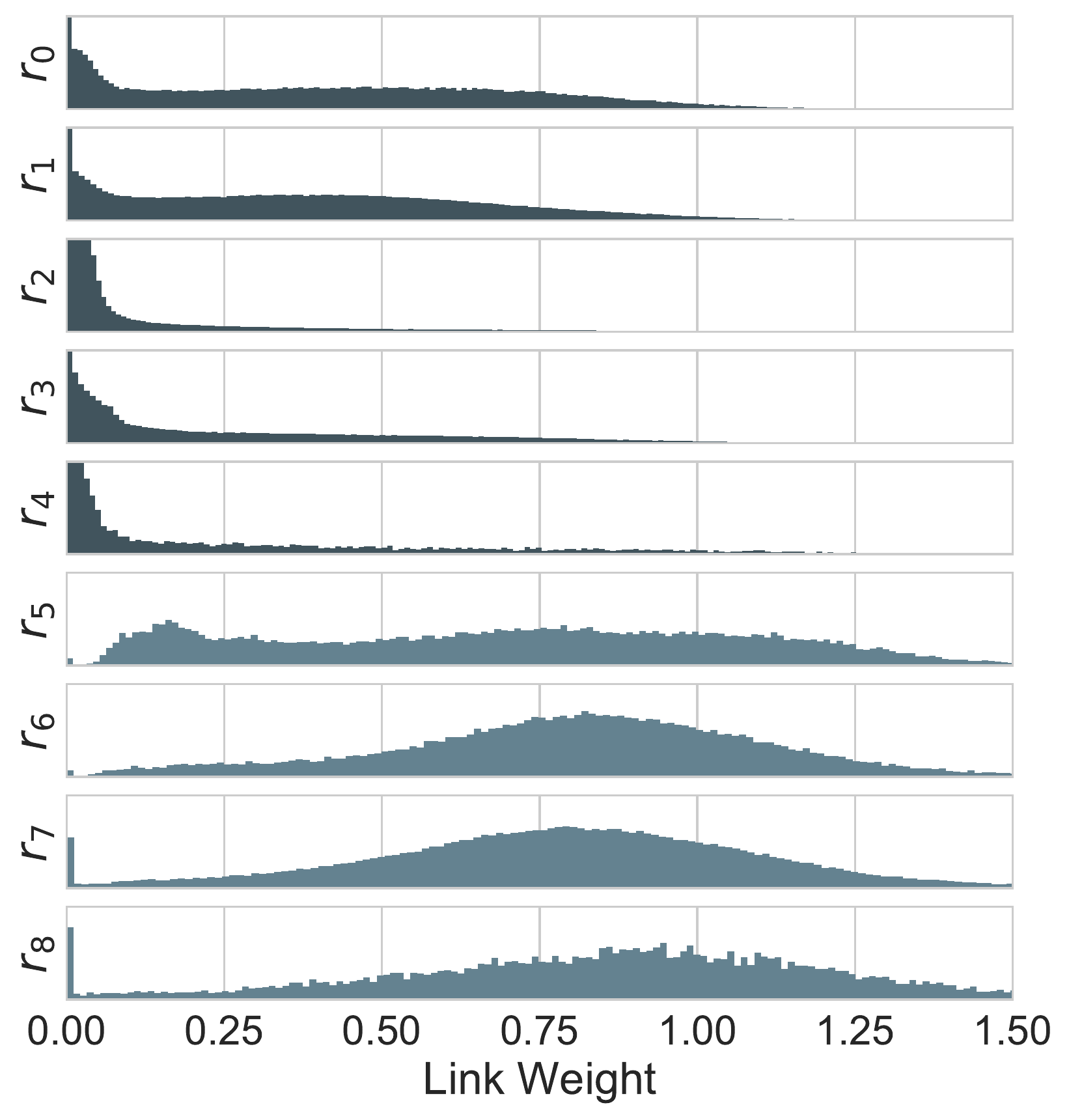}
        \label{fig:relation_distrib}
    \end{subfigure}
    \caption{
    \textbf{Left}: Ranking of a node's influencers.  The top 6 and bottom 6 known weighted-edges (+/- standard error) for cystic fibrosis are visualized as an example.
    \textbf{Centre}: Analyzing the drivers of link prediction, evaluating the possibility of PPARG being a drug target for Rheumatoid Arthritis.  Each bar demonstrates the effect that fact has on prediction score (+/- standard error).
    \textbf{Right}: Distribution of edge weights across $r \in \mathcal{R}$ in the biomedical knowledge graph.
    }
    \label{fig:viz-all}
\end{figure}

Link weights in a GCNN with attention enable the visualization of the most and least important factors underlying a representation. The left panel of Fig.~\ref{fig:viz-all} visualizes the connection weights for the disease cystic fibrosis (CF).  The strongest drivers of CF's representation are drugs listed as a treatment for CF in curated databases, supporting their relevance to the task of predicting therapies.  Two of the top six are drugs specific to CF management (Denufosol and Ivacaftor), while the other four are antibiotic drugs often used to manage infections arising in CF \cite{drugbank}.  The six lowest-weighted links, on the other hand, consist of links arising from false extractions or weak scientific evidence \cite{opentargets,ASPAgene,Lateef2005}. Furthermore, connections with ABCC6 and ABCA12 extracted from text are actually false: cystic fibrosis (CF) and these genes are merely mentioned in the same sentence as part of a list but with no functional connection.  The attention weight can illuminate edges to be rectified.

In addition to quality assessment, an edge's effect on the likelihood of a link can be examined by altering $\hat{C}_{r,i,j}$, because an edge can be removed after training in GCNNs.  This method bears similarity to feature occlusion methods used for interpretability in high-dimensional CNNs \cite{ancona2018towards,zintgraf2017visualizing}.
An example is shown in Fig.~\ref{fig:viz-all}, in which all inputs to both the left and right of a relation are independently removed while measuring the effect on the score of a therapeutic relation between the gene PPARG and the disease Rheumatoid Arthritis (RA) -- a true edge that has been hidden during training.  The top positive driver, a coexpression edge between PPARG and E2F4, uses a target in the same family as a gene implicated in RA \cite{E2F2}.  The strongest negative driver, on the other hand, is a therapeutic link between RA and PPP3CC, a gene target associated with the rather different disease schizophrenia \cite{OMIM}.  Future work could examine transductive methods of this analysis.

Finally, the usefulness of entire data sources can be assessed as a whole. The right panel of Fig.~\ref{fig:viz-all} shows the distributions of learned weights for each relation type.  The edge histogram in light grey corresponds to edge types that were trained upon (and therefore typically receive a higher learned weight), while the dark grey histograms correspond to edges present only in the adjacency matrix.  Patterns of weight distribution separate data sources.  For example, $r_1$ and $r_2$ contain proportionally more high-weighted edges than the three below, implying that these data sources are comparatively more useful for the model.  Such information enables identification of good relations and sources.

\section{Conclusion}
This work introduces an improvement to graph convolutional neural networks by adding an attention parameter for the network to learn how much to trust an edge during training.  As a result, noisier, cheaper data can be effectively leveraged for more accurate predictions.  Further, this facilitates new methods for visualization and interpretation, including ranking the influencers of a node, inferring the greatest drivers of a link prediction, and uncovering errors present in input data sources.

{\small
\bibliography{main}
\bibliographystyle{plain}
}

\beginsupplement
\section{}
\subsection{Synthetic Data Experiments: Duplication-Divergence Model}

\begin{figure}[t]
\centering
    \includegraphics[width=0.75\textwidth]{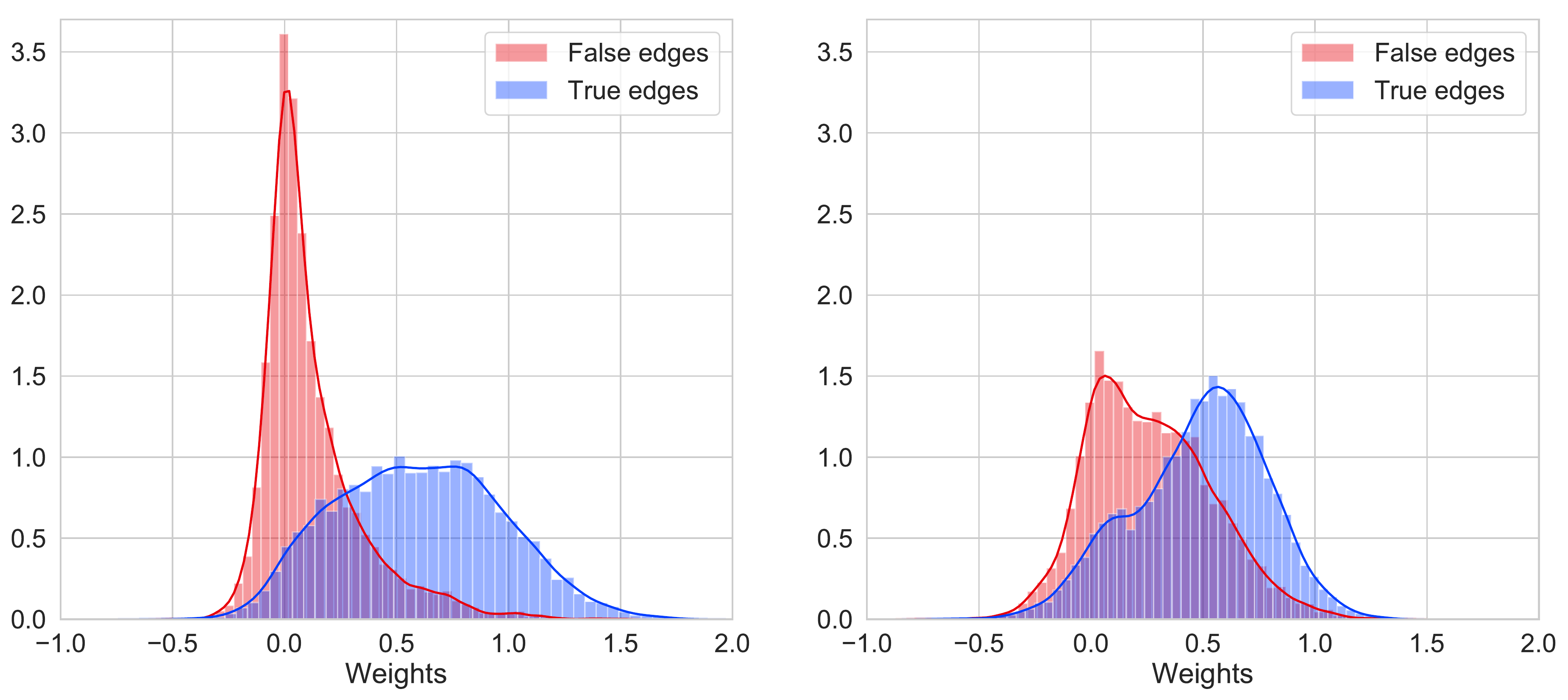}
    \caption{
    \textbf{Left}: Normalized histogram of link attention weights for edges in $\gadd$ (blue) and $\gnoise$ (red) when training only on $\ggold$.
    \textbf{Right}: Normalized histogram of link attention weights when training on edges in $\ggold \cup \gadd \cup \gnoise$.}

    \label{fig:synth_perf1}
\end{figure}

We also examined the specific advantages our model confers on synthetic data in a controlled setting, allowing perfect identification of true and false edges. To this end, we employ the Duplication-Divergence model, as commonly used to model protein-protein interaction networks \citep{Ispolatov2005-id}.

The Duplication-Divergence (DD) Model \cite{Ispolatov2005-id} is a two-parameter model $p, q\in [0, 1]$ that mimics the growth and evolution of protein-protein interaction networks. Given a small starting seed graph $\mathcal{G}_0$ with genes as nodes and edges representing an interaction between their encoded proteins, the model simulates the process of gene duplication and mutation. To execute this model, first uniformly select a node $n_\text{old}$ in $\mathcal{G}_0$ and define a new node $n_\text{new}$. Next, connect $n_\text{new}$ and $n_\text{old}$ with probability $q$ and connect $n_\text{new}$ with any other node linked to node $n_\text{old}$, each with probability $p$. This results in a network $\mathcal{G}_1$, and the process is repeated until the network grows to a specified size.


Let $\mathcal{G}_0$ be a graph of two connected nodes. We define a 1000-node graph $\mathcal{G}_{1000}$ with parameters $p=0.75, q=0.0$. The value of $p$ and $q$ are chosen to ensure that each of the sub-sampled training graphs has a single connected component with edge-vertex ratio $\sim 20$, reflecting the estimated size of the human interactome \cite{Stumpf2008-um}.

This setup recreates the condition of an incomplete but high-precision dataset ($\ggold$) with which to train a model for link-prediction, together with the option of supplementing the training data with a set of possibly noisy relations ($\gadd \cup \gnoise$).  The model here is a single-layer GCNN, with embedding size 300, and a diagonalized weight matrix $W_r$ is chosen with no non-linearity transformation applied.  Fig.\ref{fig:synth_perf1}, left two plots, demonstrate that a low attention weight can imply a faulty edge, and that the identification of false edges works best when only training on high-quality edges ($\ggold$) while the adjacency matrix $A_{r,i,j}$ can include edges from a noiser set ($\gadd \cup \gnoise$).

\end{document}